\def\year{2021}\relax
\documentclass[letterpaper]{article} % DO NOT CHANGE THIS
\usepackage{aaai21}  % DO NOT CHANGE THIS
\usepackage{times}  % DO NOT CHANGE THIS
\usepackage{helvet} % DO NOT CHANGE THIS
\usepackage{courier}  % DO NOT CHANGE THIS
\usepackage[hyphens]{url}  % DO NOT CHANGE THIS
\usepackage{graphicx} % DO NOT CHANGE THIS
\usepackage{natbib}  % DO NOT CHANGE THIS AND DO NOT ADD ANY OPTIONS TO IT
\usepackage{caption} % DO NOT CHANGE THIS AND DO NOT ADD ANY OPTIONS TO IT
\usepackage[utf8]{inputenc} % allow utf-8 input
\usepackage[T1]{fontenc}    % use 8-bit T1 fonts
\usepackage{booktabs}       % professional-quality tables
\usepackage{amsfonts}       % blackboard math symbols
\usepackage{nicefrac}       % compact symbols for 1/2, etc.
\usepackage{microtype}      % microtypography
\usepackage{color}

\usepackage{algorithm,algcompatible,amsmath}
\usepackage{algpseudocode} 
\usepackage{multicol}
\usepackage{multirow}
\usepackage{graphicx}
\usepackage[caption=false,font=footnotesize]{subfig}
\newcommand{\deflen}[2]{%      
    \expandafter\newlength\csname #1\endcsname
    \expandafter\setlength\csname #1\endcsname{#2}%
}

\usepackage[switch]{lineno}

\title{Gradient Regularized Contrastive Learning for Continual Domain Adaptation}

\author{
        Shixiang Tang\footnote{denotes equal contribution},\textsuperscript{\rm 1 3}
        Peng Su$^{*}$, \textsuperscript{\rm 2}
        Dapeng Chen, \textsuperscript{\rm 3} 
        Wanli Ouyang \textsuperscript{\rm 1}\\
}
\affiliations{
    \textsuperscript{\rm 1} University of Sydney, Australia \\
    \textsuperscript{\rm 2} The Chinese University of Hong Kong, Hong Kong \\
    \textsuperscript{\rm 3} Sensetime Group Limited, Hong Kong \\
    \{stan3906,wanli.ouyang\}@uni.sydney.edu.au, psu@ee.cuhk.edu.hk, chendapeng@sensetime.com
}

\begin{document}
% \linenumbers
\maketitle
\begin{abstract}
Human beings can quickly adapt to
environmental changes by leveraging learning experience. 
However, adapting deep neural networks to dynamic environments by machine learning algorithms remains a challenge.
To better understand this issue, we study the problem of continual domain adaptation, where the model is presented with a labelled source domain and a sequence of unlabelled target domains. The obstacles in this problem are both domain shift and catastrophic forgetting. We propose Gradient Regularized Contrastive Learning (GRCL) to solve the obstacles.
At the core of our method, gradient regularization plays two key roles: (1) enforcing the gradient not to harm the discriminative ability of source features which can, in turn, benefit the adaptation ability of the model to target domains; (2) constraining the gradient not to increase the classification loss on old target domains, which enables the model to preserve the performance on old target domains when adapting to an in-coming target domain. Experiments on Digits, DomainNet and Office-Caltech benchmarks demonstrate the strong performance of our approach when compared to the other state-of-the-art methods.
\end{abstract}

\section{Introduction}
Generalizing models learned from one domain (source domain) to novel domains (target domains) has been a major challenge of machine learning.  The performance of the model learned on one domain may degrade significantly on other domains because of different data distribution~\cite{luo2019taking,ben2010theory,moreno2012unifying,storkey2009training,yu2020cocas}. In this work, we investigate continual domain adaptation (continual DA), where models are trained in multi-steps and only part of training samples are presented in each step. 
Continual DA considers the real-world setting where target domain data are acquired sequentially. As an example, autonomous driving requires adapting to scenes (target domains) in different weathers and different countries. And these training data are usually collected in different seasons: snowy scenes can only be collected in winter while rainy scenes are mostly in summer. Continual DA also considers efficiency when the model is deployed in the real world. When samples from a novel domain are acquired, conventional DA, i.e. not continual DA, requires to use all samples collected to train the model from scratch, which is time-consuming. Continual DA can address the problem as it enables the model incrementally adapts to the new domain without losing generalization ability on old domains.

In this paper, we consider keeping the discriminative ability of source features can benefit adapting the model to different target domains. Intuitively, as the labels of source domain samples are given, the discriminative ability learned from labelled source samples can guide the adaptation to all target domains.  Existing DA methods~\cite{ganin2016domain,long2017deep,tzeng2017adversarial,saito2018maximum,su2020adapting,peng2019domain} cannot retrain such discriminative ability on source features because they adopt multitask learning, \emph{i.e.} one classification loss on the source domain and one domain adaptation loss. When minimizing the multitask loss, the classification loss on the source domain may still increase, meaning the model's discriminative ability on the source domain is weakened. In contrast, our method constrains the classification loss on the source domain non-increase (\emph{source discriminative constraint}) in every training iteration. In this way, we maintain the discriminative ability of source features and more importantly improve the adaptation ability of the learned model to the target domain.

Furthermore, the model trained with sequential data suffers from \emph{catastrophic forgetting}. Existing method~\cite{bobu2018adapting} handles \emph{catastrophic forgetting} by incorporating a replay in the adversary training framework. However, this approach assumes that the sequential target domain shift follows some specific patterns and suffers from \emph{catastrophic forgetting} when the assumption breaks. In contrast, when adapting the model to a new target domain, we propose to enforce the classification loss not to increase for every old target domain (\emph{target memorization constraint}).  This constraint ensures the model not to lose the generalization ability on old target domains when adapting to a new target domain.

Based on the observations above, we propose gradient regularized contrastive learning (GRCL), to tackle continual DA. 
GRCL leverages the contrastive loss to learn domain-invariant representations using the samples in the source domain, the old target domains and the new target domain. Two constraints, i.e. \emph{source discriminative constraint} and \emph{target memorization constraint}, are proposed when optimizing the network. Specifically, the \emph{source discriminative constraint} is formulated to constrain that the gradient of the parameters should be positively correlated to the gradient of classification loss for the source domain. And the \emph{target memorization constraint} constrains that the gradient of the parameters should be positively correlated to the gradient of classification loss for every old target domain. The pseudo-labels involved in the \emph{target memorization constraint} are generated by clustering\cite{Yang_2020_CVPR,Yang_2019_CVPR,Guo_2020_CVPR,ester1996density,1017616}. These pseudo-labels are high-quality because features in old target domains are discriminative and we can filter out those samples with low confidence.

To summarize, our contributions are as follows:
(1) We propose a \emph{source discriminative constraint} to improve discriminative ability of features in the target domains by preserving the discriminative ability of source features.
(2) We propose a \emph{target memorization constraint} to explicitly memorize the knowledge on old target domains. The proposed two constraints consistently improve
the continual DA by over $5\%$ compared with the baseline method on three benchmarks.

\section{Related Works}
\subsection{Unsupervised Domain Adaptation} 
Unsupervised domain adaptation (UDA) aims to transfer the knowledge from a different but related domain (source domain) to a novel domain (target domain). Various methods have been proposed, including discrpancy-based UDA approaches~\cite{long2017deep,tzeng2014deep,ghifary2014domain,peng2018synthetic}, adversary-based approaches~\cite{liu2016coupled,tzeng2017adversarial,liu2018unified,su2020adapting}, reconstruction-base approaches~\cite{yi2017dualgan,zhu2017unpaired,hoffman2018cycada,kim2017learning} and contrastive learning based approaches~\cite{ge2020selfpaced,park2020joint,kim2020cross}. 
To maintain the discriminative ability of the model to the source domain, these approaches resort to adding the task loss, e.g. classification loss for the image classification task, on the source domain data to the adaptation loss as a multitask objective function. Consequently, the task loss on the source domain data often increases even though the multitask objective function is minimized.
In contrast, we explicitly enforce that the task loss on the source domain should not increase, preserving the discriminative ability of the model to the source domain.
Besides, instead of setting the trade-off parameter manually in the multi-task learning, our GRCL adaptively updates the ratio of gradients in different tasks by solving an optimization problem. Such adaptive updating of the gradient is the key to maintaining the discriminative ability of source features, which, in turn, improves performance on the target domain.

\subsection{Continual Learning} Continual learning~\cite{prabhu2020greedy,PARISI201954,zhao2020continual} addresses \emph{catastrophic forgetting} in a sequence of supervised learning tasks. Popular methods can be categorized as regularization-based methods~\cite{kirkpatrick2017overcoming,zenke2017continual,aljundi2018memory,li2017learning} and memory-based methods~\cite{rebuffi2017icarl,castro2018end,hou2018lifelong,hou2019learning}.
In particular, GEM\cite{lopez2017gradient} is most related to our method. It uses episodic memory to store some training samples of old tasks and conducts constrained optimization to address the \emph{catastrophic forgetting} problem.  However, 
GEM cannot address continual DA very well, it aims to learn different classes continuously, while continual DA needs to recognize the images with the same label space but from different domains. 
To address the continual DA, our method highlights the importance of the source domain data and explicitly utilizes the constraints from the source domain to guide the learning of all other target domains.

\subsection{Continual Domain Adaptation}
When learning a sequence of unlabeled target domains, continual domain adaptation aims to achieve good generalization abilities on all seen domains~\cite{lao2020continuous}. Massimiliano~\cite{mancini2019adagraph} attempted to solve a specific scenario in continual DA, where no target data is available, but with metadata provided for all domains. Gong~\cite{gong2019dlow} proposed to bridge two domains by generating a continuous flow of intermediate states between two original domains. Several other papers~\cite{wulfmeier2018incremental,hoffman2014lsda,wang2020continuously,cheung2019superposition} presented continuous domain adaptation with the emphasis to generalize on a transitioning target domain. The approaches \cite{mancini2019adagraph,gong2019dlow,wulfmeier2018incremental,hoffman2014lsda} do not explicitly target the \emph{catastrophic forgetting} problem, while the approaches \cite{bobu2018adapting} and our approach target the \emph{catastrophic forgetting} problem.
Our approach is different from the method in two aspects.
First, existing works aim to address \emph{catastrophic forgetting}, but with an implicit assumption that the domain shift follows a specific pattern, \emph{i.e.} data shift domain gradually, e.g. gradually changing weather or lighting condition. However, our approach does not need this assumption because we explicitly set constraints on every old target domain without relying on the domain relationships.
%Second, we focus on more general use cases for solving any arriving domain without imposing \wl{extra constraints on the relationships among the domains}.
Second, compared with the closest work in~\cite{bobu2018adapting} which uses multitask learning with a simple replay, our GRCL
emphasizes the importance of the discriminative source features and tackles \emph{catastrophic forgetting} by strictly following the constraint that the task loss on the old target domains non-increase in every iteration.

\begin{figure*}[]
    \centering
    \includegraphics[width=1\linewidth]{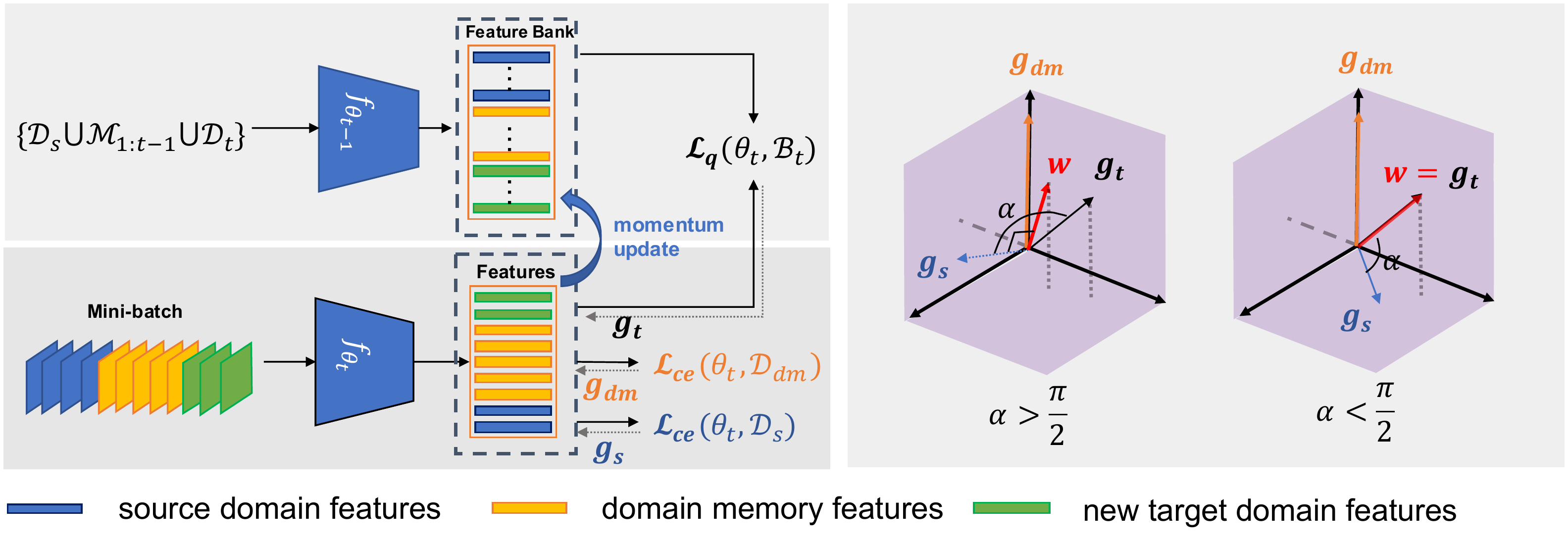}
    \caption{Gradient Regularized Contrastive Learning. Left: Schematic representation of GRCL. The  feature bank is initialized by pretrained model $f_{\theta_{t-1}}$. During training, mini-batch samples are sampled from the source domain, domain-episodic memories and the new target domain. Contrastive loss, \emph{source discriminative constraint}, \emph{target memorization constraint} are then imposed on the mini-batch. The features of the mini-batch are involved in updating the feature bank in a momentum way. Right: The gradient regularization is utilized to enforce the  gradient of contrastive loss not to increase the cross-entropy loss, which maintains the discriminative ability of feature learned by contrastive learning.}
    \label{fig:fig2}
\end{figure*}

\section{Problem Formulation}
Let $\mathcal{D}_s = \{ (x_i^s, y_i^s) \}_{i=1}^{n_s}$ be the labeled dataset of source domain, where each example $(x_i^s, y_i^s)$ is composed of an image $x_i^s \in \mathcal{X}^s$ and a label $y_i^s \in \mathcal{Y}$.
Continual domain adaptation defines a sequence of adaptation tasks $\mathcal{T}_{1:N} = \{ \mathcal{T}_1, \mathcal{T}_2, \dots, \mathcal{T}_N \}$.
On $t$-th task $\mathcal{T}_t$, there is an {\it unlabeled} target domain dataset $\mathcal{D}_t =\{ x_i^t\}_{i=1}^{n_t}$. Different domains share a common label space $\mathcal{Y}$ but have distinct data distributions.
The goal is to learn a label prediction model $f$ that can generalize well on multiple target domains $\{\mathcal{D}_1, \dots, \mathcal{D}_N \}$. 

We propose two metrics to evaluate the model adapting over a stream of target domains, namely average accuracy (ACC) and average backward transfer (BWT).
After the model adapts to the target domain $\mathcal{D}_t$, we evaluate its performance on the testing set of the new and all old target domains $\mathcal{D}_k^{test} (\forall k \leq t)$.
Let $R_{t, j}$ denote the test accuracy of the model on the domain $\mathcal{D}_j$ after adapting the model to domain $\mathcal{D}_t$. We use $\mathcal{D}_0$  to denote the source domain.
ACC and BWT can be calculated as
\begin{equation}
     \small
     \text{ACC}  = \frac{1}{N} \sum_{t=0}^{N} R_{N, t},~\text{BWT}  = \frac{1}{N-1} \sum_{t=1}^{N-1} R_{N, t} - R_{t, t}.
\end{equation}
The ACC represents the average performance over all domains when the model finishes all sequential adaptation tasks. BWT indicates the influence on previously observed domains $\mathcal{D}_{k < t}$ when adapting to domain $\mathcal{D}_t$.  
The negative BWT indicates that adapting to a new domain decreases the performance on previous domains. The larger these two metrics are, the better the model is.

\section{Methodology}
We propose a gradient regularized contrastive learning framework (GRCL) to tackle the challenges in continual unsupervised domain adaptation (Continual DA). When adapting the model to the $t$-th target domain, the baseline framework is based on contrastive learning with domain-episodic memories $\mathcal{M}_{1:t-1}$ and a feature bank $\mathcal{B}_t$. The key innovation of GRCL lies in two novel constraints on gradients when jointly training samples in source domain $\mathcal{D}_s$, domain-episodic memories $\mathcal{M}_{1:t-1}$ and the new target domain $\mathcal{D}_t$ in a contrastive way. The \emph{source discriminative constraint} can maintain the discriminative ability of samples in the source domain and surprisingly, in turn, improves the adaptability on the target domain.  The \emph{target memorization constraint} overcomes \emph{catastrophic forgetting} on old target domains when adapting the model to a new target domain.

In greater detail, the training samples in each batch are sampled from the source domain~$\mathcal{D}_s$, episodic memories~$\mathcal{M}_{1:t-1}$ and the new target domain~$\mathcal{D}_t$. These samples are trained with the unified contrastive loss~(Eq.~\ref{eq:2}) with the help of the feature bank $\mathcal{B}_t$ by Eq.~\ref{eq:bank_init}. \emph{Source discriminative constraint}~(Eq.~\ref{eq:4}) and \emph{target memorization constraint}~(Eq.~\ref{eq:source-constraint}) are imposed on samples of the source domain and the episodic memories in the minibatch respectively. The gradient to update the model $w$ is computed by solving the quadratic optimization problem (Eq.~\ref{eq:8}). The whole pipeline is illustrated in Fig.~\ref{fig:fig2}(left).

\subsection{Baseline Framework with Contrastive Learning}
Contrastive learning~\cite{wu2018unsupervised,he2019momentum,chen2020simple} has recently shown the great capability of mapping images to an embedding space, where similar images are close together and dissimilar images are far apart. Inspired by this, we utilize the contrastive loss to push the target instance towards the source instances that have similar appearances with the target input. To better exploit the features from the source domain, old target domains and the new target domain, we unify these features in one feature bank and introduce a unified contrastive loss with the feature bank in detail.

\subsubsection{Feature Bank} \quad We propose a  feature bank $\mathcal{B}_t$ to provide source features, representative old target domain features and new target domain features. We initialize the feature bank as
$\mathcal{B}_t=\{ k(x), \forall x \in  \mathcal{D}_s \cup \mathcal{M}_{1:t-1}  \cup \mathcal{D}_t \}$, where $\mathcal{M}_{1:t-1}=\{\mathcal{M}_1, \mathcal{M}_2, ..., \mathcal{M}_{t-1}\}$ and $\mathcal{M}_i$ stores representitive samples in the target domain $\mathcal{D}_i$. In particular, $k(x)$ is a  representation of the input $x$ and can be computed by 
\begin{equation} \label{eq:bank_init}
    k(x) = q_{t-1}(f_{\theta_{t-1}}(x)),
\end{equation}
where $f_{\theta_{t-1}}$ is a CNN-based encoder, and $q_{t-1}$ is a MLP projector after adapting the model to $(t-1)$-th target domain. All features are normalized by $\| k(x) \|_2^2 =1$. At each training iteration, the encoded features $q_t(f_{\theta_t}(x))$ in the mini-batch will be used to update the memory bank $\mathcal{B}_t$ by the rule: 
\begin{equation} \label{eq:momentum_update}
k(x) \leftarrow mk(x)+(1-m)q_t(f_{\theta_t}(x)), \quad m\in [0,1]
\end{equation}

\subsubsection{Unified Constrastive Loss} \quad  For adapting the model to the $t$-th target domain, the CNN-based $f_{\theta_{t}}$ is initialized by $f_{\theta_{t-1}}$, the MLP projection $q_t$ is initialized by $q_{t-1}$ and the feature bank $\mathcal{B}_t$ is initialized by Eq.~\ref{eq:bank_init}. Images in each training batch are sampled from the source domain $\mathcal{D}_s$, the episodic memories $\mathcal{M}_{1:t-1}$ and the new target domain $\mathcal{D}_t$ at a fixed ratio. The contrastive loss \cite{oord2018representation} computed by each training batch is: 
\begin{equation} \label{eq:2}
\mathcal{L}_{\mathbf{q}}(\theta_t, \mathcal{B}_t) = -\log \frac{\exp (\mathbf{q} \cdot \mathbf{k}^+ / \tau)}{\exp (\mathbf{q} \cdot \mathbf{k}^+ / \tau) + \sum\limits_{\mathbf{k}^{-}\in   \mathcal{B}_t} \exp(\mathbf{q} \cdot \mathbf{k}^{-} / \tau)}, 
\end{equation}
where $\mathbf{q}$ is a general feature vector $\mathbf{q}=q_t(f_{\theta_t}(x))$ and $x$ denotes the samples in the training batch. $\mathbf{k}^{+}$ is the positive key for $\mathbf{q}$ and can be defined as the corresponding feature of sample $x$ stored in $\mathcal{B}_t$. Features other than $\mathbf{k}^+$ in $\mathcal{B}_t$ can be used as negative keys $\mathbf{k}^{-}$ for $\mathbf{q}$. The temperature $\tau$ is empirically set as 0.07. 

\begin{figure*}[t]
    \centering
    \includegraphics[width=1\linewidth]{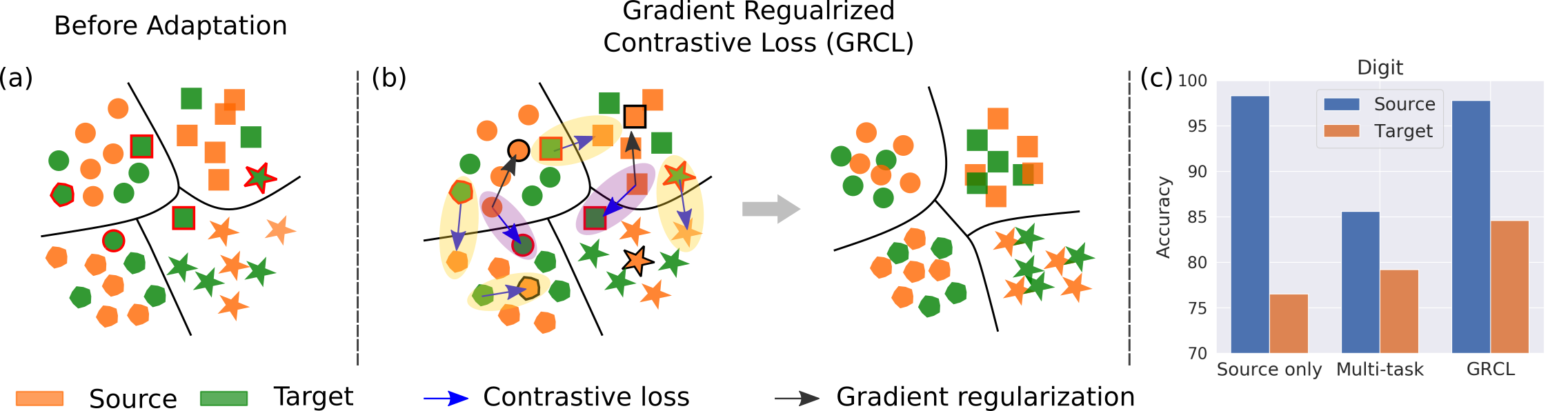}
    \caption{Illustration of GRCL. In Fig.(a,b) the shape denotes class and color denotes domain.
    The contrastive loss pushes target samples towards its similar ones in the source domain. 
    However, it inevitably pushes some discriminative features towards some less discriminative features (purple area), as the contrastive loss only attracts the visually similar features.
    Such a problem can be verified by the severe performance degradation on the source domain (Fig.(c)).
    GRCL regularizes the model update not to increase the loss on source domain (black arrow), that maintains the discriminative ability of learned features.
    }
    \label{fig:demo-grcl}
\end{figure*}

\subsection{Source Discriminative Constraint}
The contrastive loss can bridge the domain gap by attracting visually similar samples, but it may harm the discriminative ability of features in the source domain. Examples are shown in ~Fig.~\ref{fig:demo-grcl}(b). The orange squares/circles in the purple area are pulled towards their green counterparts, hence the discriminative ability of source features is deteriorated. As labels in the source are reliable, the knowledge learnt from the source domain are valuable for all the target domains. We therefore add the \emph{source discriminative constraint} when minimizing the contrastive loss Eq.~\ref{eq:2}, which is to regularize the $\mathcal{L}_{ce}(\theta_t, \mathcal{D}_s)$ on the source domain non-increase: 
\begin{equation} \label{eq:source-constraint}
     \mathcal{L}_{ce}(\theta_t, \mathcal{D}_s) \leq \mathcal{L}_{ce}(\theta_{t-1}, \mathcal{D}_s).
\end{equation}
We denote $w$ as the vector to update the model and $g_s$ as the gradient of $\mathcal{L}_{ce}(\theta_t, \mathcal{D}_s)$. Inspired by \cite{lopez2017gradient}, the Eq.~\ref{eq:source-constraint} can be rephrased as
\begin{equation} \label{eq:source-constraint-grad}
     \langle w,  g_s \rangle \geq 0 ,
\end{equation}
where $\langle w,  g_s \rangle$ means the inner product between the update vector and the original gradient of the classification loss on the source domain. 
\begin{equation} \label{eq:4}
 \langle w, g_{s} \rangle := \langle w, \frac{\partial \mathcal{L}_{ce}(\theta_t, \mathcal{D}_s)}{ \partial \theta_t} \rangle \geq 0.
\end{equation}

As illustrated in Fig.\ref{fig:fig2} (right),  if the angle between $g_s$ and $g_t$ is less than $\pi/2$, minimizing Eq.~\ref{eq:2}  by $g_t$ will not increase the classification loss on the source domain. Therefore, we simply use $w=g_t$ to update the model parameters. 
If the angle between $g_s$ and $g_t$ is larger than $\pi/2$, updating parameters by $g_t$ will inevitably increase the classification loss the source domain. 
We therefore enforce $w$ to satisfy Eq. \ref{eq:4} while
keeping close to the gradient of the contrastive loss $g_{t}$. 

% In this case, we propose to discompose $g_t$ in the plane defined by $g_s$ and $g_t$. We compute $w$ as the perpendicular component of $g_t$ w.r.t $g_s$. The perpendicular relation between $w$ and $g_s$ can avoid increasing $\mathcal{L}_{ce}(\theta_t, \mathcal{D}_s)$ while sacrificing the efficiency to minimize Eq.~\ref{eq:2}.

\subsection{Target Memorization Constraint}
An essential problem in continual domain adaptation is \emph{catastrophic forgetting}. For this problem, \emph{Target Memorization Constraint} is proposed to keep the 
classification loss for each domain-episodic memory non-increase:
% To overcome it, we construct a constraint that the classification loss for each domain-episodic memory $\mathcal{M}_i$ never increases: 
\begin{equation} \label{eq:6-0}
 \mathcal{L}_{ce}(\theta_t, \mathcal{M}_i) \leq \mathcal{L}_{ce}(\theta_{t-1}, \mathcal{M}_i )\quad \text{for all} \quad i < t,
\end{equation}
where $\mathcal{M}_i$ is $i$-th domain-episodic memory. When the number of target domains increases, the computation burden for Eq.~\ref{eq:6-0} will be large, because we need to solve the $t\!-\!1$ constraints of all domain-episodic memory. 
Alternatively, we leverage a much efficient way to approximate the Eq.\ref{eq:6-0} by

\begin{equation} \label{eq:6}
 \mathcal{L}_{ce}(\theta_t, \mathcal{M}_{1:t-1}) \leq \mathcal{L}_{ce}(\theta_{t-1}, \mathcal{M}_{1:t-1} ),
\end{equation}
where $\mathcal{M}_{1:t-1} = \cup_{i=1}^{t-1} \mathcal{M}_i $.
Instead of computing the loss on each individual previous domain, Eq.~\ref{eq:6} only computes the loss with the sampled batch of images from $\mathcal{M}_{1:t-1}$. Similarly, Eq.~\ref{eq:6} can be rephrased as
\begin{equation} \label{eq:target_gradient}
 \langle w, g_{dm} \rangle := \langle w, \frac{\partial \mathcal{L}_{ce}(\theta_t, \mathcal{M}_{1:t-1})}{ \partial \theta_t} \rangle \geq 0.
\end{equation}

\subsection{Overall Formulation and Solution of GRCL}
We combine contrastive learning (Eq.~\ref{eq:2}) with \emph{source discriminative constraint} (Eq.~\ref{eq:4}) and \emph{target memorization constraint} (Eq.~\ref{eq:target_gradient}) and then propose the overall objective function (Eq.~\ref{eq:8}) to obtain the final parameter update vector.

In order to incorporate gradient regularization in contrastive loss minimization, we modify the objective $\text{min}~ \mathcal{L}_{\textbf{q}}$ to $\text{min}~||w-g_t||^2_2$ for each iteration, where $g_t$ is the gradient of contrastive loss and $w$ is the gradient to update the network. The rationality behind is that to efficiently minimize contrastive loss, $w$ need to be as close to $g_t$ as possible under the constrains Eq.~\ref{eq:4} and Eq.~\ref{eq:6}. Mathematically, the overall formulation for GRCL can be defined as
\begin{equation}   \label{eq:8}
    \begin{aligned}
    \min_{w} \quad & \frac{1}{2} \| w -  g_t \|_2^2 \\
    \text{subject to} \quad  & \langle w,  g_s \rangle \geq 0 \\
     & \langle w, g_{dm} \rangle \geq 0.
    \end{aligned}
\end{equation}
Eq.~\ref{eq:8} is essentially a quadratic programming (QP) problem.  Directly solving this problem will involve a huge number of parameters (the number of parameters in the neural network). To solve the Eq.\ref{eq:8} efficiently, we work in the dual space,
resulting in much smaller QP with only $2$ variables:
\begin{equation} \label{eq:10}
    \min_{u} \quad   \frac{1}{2} u^{\top}GG^{\top}u + g_t^{\top}G^{\top}u  \quad
    \text{subject to} \quad   u \geq 0,
\end{equation}
where $G = -(g_s, g_{dm}) \in \mathbf{R}^{ 2 \times P}$ and we discard the constant term of $g_t^{\top}g_t $.
The formal proof of Eq.\ref{eq:10} is provided in Supplemetry Materials.
Once the solution $u^*=(u^*_1, u^*_2)$ to Eq.\ref{eq:10} is found, we can solve the  Eq.\ref{eq:8} by $w = G^{\top}u^* + g_t=g_t-u^*_1g_s-u^*_2 g_{dm}$.
The training protocol of GRCL is summarized in Supplementary Materials.

\begin{table*}[!h]
\centering
\begin{tabular}{@{}lcc|cc|cc@{}}
\toprule
Methods & \multicolumn{2}{c}{\textit{Digits}}                                                                       & \multicolumn{2}{c}{\textit{DomainNet}}                                     & \multicolumn{2}{c}{\textit{Office-Caltech}} \\ \midrule
                                    & ACC                               & BWT                                    & ACC                                  & BWT                                    & ACC                                   & BWT          \\ \cmidrule(l){2-3} \cmidrule(l){4-5} \cmidrule(l){6-7} 

DANN~\cite{ganin2016domain}         & 74.56 \scriptsize $\pm$ 0.14      & -11.37 \scriptsize $\pm$ 0.09          & 30.18  \scriptsize $\pm$ 0.13        & -10.27  \scriptsize $\pm$ 0.07          & 81.78 \scriptsize $\pm$ 0.05        & -8.75    \scriptsize $\pm$ 0.07         \\
MCD~\cite{saito2018maximum}         & 76.46 \scriptsize $\pm$ 0.24       & -10.90 \scriptsize $\pm$ 0.11          & 31.68 \scriptsize $\pm$ 0.20        & -10.36  \scriptsize $\pm$ 0.15           & 82.63 \scriptsize $\pm$ 0.13        & -8.70   \scriptsize $\pm$ 0.12           \\
DADA~\cite{peng2019domain}          & 77.30 \scriptsize $\pm$ 0.19       & -11.40 \scriptsize $\pm$ 0.04          & 32.14  \scriptsize $\pm$ 0.14       & -8.67  \scriptsize $\pm$ 0.09           & 82.05 \scriptsize $\pm$ 0.03        & -8.30   \scriptsize $\pm$ 0.05           \\ 
CUA~\cite{bobu2018adapting}         & 82.12 \scriptsize $\pm$ 0.18        & -6.10 \scriptsize $\pm$ 0.12          & 34.22  \scriptsize $\pm$ 0.16       & -5.53  \scriptsize $\pm$ 0.14          & 84.83  \scriptsize $\pm$ 0.10       & -4.65   \scriptsize $\pm$ 0.08           \\ \hline
GRA                               & 84.10 \scriptsize $\pm$ 0.15        & -0.93 \scriptsize $\pm$ 0.10          & 35.84  \scriptsize $\pm$ 0.19       & -1.15  \scriptsize $\pm$ 0.16           & 86.53  \scriptsize $\pm$ 0.11       & -0.03   \scriptsize $\pm$ 0.03           \\
GRCL                                 & 85.34 \scriptsize $\pm$ 0.10        & -1.0 \scriptsize $\pm$ 0.03          & 37.74 \scriptsize $\pm$ 0.13        & -0.67  \scriptsize $\pm$ 0.12           & 87.23  \scriptsize $\pm$ 0.06       & 0.05      \scriptsize $\pm$ 0.02            \\ \bottomrule
\end{tabular}
\caption{ACC and BWT on three continue domain adaptation benchmarks.}
\label{tab:t1}
\end{table*}

\subsubsection{Discussion} 

In the context of our paper, the previous methods solve DA or continual DA as a multitask learning problem. For example, the loss function of continual DA can be formulated as $\mathcal{L} = \mathcal{L}_{\mathbf{q}}(\theta_t, \mathcal{B}_t) + \lambda_1 \mathcal{L}_{ce}(\theta_t, \mathcal{D}_s) + \lambda_2 \mathcal{L}_{ce}(\theta_t, \mathcal{M})$,
% \begin{equation} \label{eq:contrast-da-multi-task}
%     \begin{aligned}
%     \min_{\theta_t}~\mathcal{L} & = \mathcal{L}_{\mathbf{q}}(\theta_t, \mathcal{B}_t) + \lambda_1 \mathcal{L}_{ce}(\theta_t, \mathcal{D}_s) + \lambda_2 \mathcal{L}_{ce}(\theta_t, \mathcal{M}),
%     \end{aligned}
% \end{equation}
where $\lambda_1, \lambda_2$ is the hyper-parameter to trade off the three losses.

Our gradient regularized method differs from multitask learning in two aspects. (1) GRCL ensures that parameters update will not harm the classification loss on both source domain and old target domains. In contrast, multitask learning only minimizes the overall loss, without \emph{source discriminative constraint} or \emph{target memorization constraint} guaranteed. (2) Trade-off parameters are different in $w=g_t + \lambda_1 g_s + \lambda_2 g_{dm}$ of multi-task learning and $w = g_t - u^*_1 g_s - u^*_2 g_{dm}$ of GRCL. Compared with $\lambda_1, \lambda_2$ that are given manually, $u^*_1, u^*_2$ are computed by Eq.~\ref{eq:10} and are adaptive in each iteration. Therefore, we conclude that GRCL can better balance the importance of different sub-losses during each iteration.

For the experiments on adapting the model to the first target domain, the results (Fig.\ref{fig:demo-grcl} (c)) show that the multitask loss only brings marginal improvements on the target domain but degrades the performance on the source domain. In contrast, our gradient regularized method can significantly improve the performance on the target while preserving the performance on the source domain.

% We compare the multitask loss with the proposed gradient-regularized method by experiments on single domain adaptation problem. The result 
% We attribute the marginal improvements to not fully preserving and exploiting the discriminative ability of the source features.

% \begin{figure}[t]
%     \centering
%     \includegraphics[width=0.6\linewidth]{Figures/gdm-sample.png}
%     \caption{Comparison between unsupervised domain adaptation method (UDA) and  Gradient Regularized Contrastive Learning (GRCL).
%     $\theta$ represents the current model position in the parameter space, and the black arrow curve denotes the direction of optimization trajectory.
%     GRCL finds the solution that minimizes the loss on a new domain $t$ while not increase the loss on the previously observed domain $k$ ($k <t$).
%     UDA tends to maximize the performance on the domain $t$ but sacrifice the performance on a preceding domain $k$.}
%     \label{fig:UDA-VS-GDM}
% \end{figure}
\begin{figure*}[h]
    \centering
    \includegraphics[width=1\linewidth]{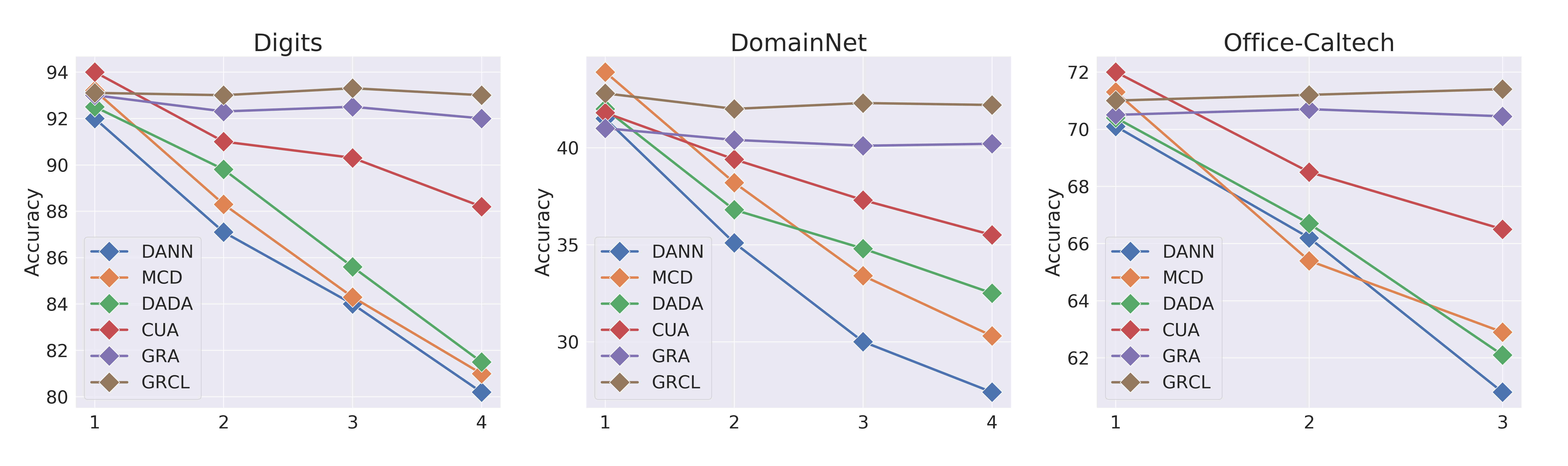}
    \caption{Evolution of classification accuracy on the first target domain as more domains are observed. Existing methods (DANN, MCD, DADA, CUA) exhibit significant performance degradation due to \emph{catastrophic forgetting}.}
    \label{fig:fig3}
\end{figure*}
\section{Experiment}
\subsection{Experimental Setup}
\subsubsection{Datasets} We test our method on three popular datasets. 

\emph{Digits} includes five digits datasets (MNIST~\cite{lecun1998gradient}, MNIST-M~\cite{ganin2015unsupervised}, USPS~\cite{hull1994database}, SynNum~\cite{ganin2015unsupervised} and SVHN~\cite{netzer2011reading}).
Each domain has $7,500$ images for training and $1,500$ images for testing.
We consider a continual domain adaptation problem of SynNum $\xrightarrow{}$ MNIST $\xrightarrow{}$ MNIST-M   $\xrightarrow{}$  USPS $\xrightarrow{}$ SVHN. \\ 
\emph{DomainNet}~\cite{Peng_2019_ICCV} is one of  the largest domain adaptation datasets with approximately $0.6$ million images distributed among $345$ categories.
Each domain randomly selects $40,000$ images for training and $8,000$ images for testing.
Five different domains from DomainNet are used to build a continual domain adaptation task as Clipart $\xrightarrow{}$ Real $\xrightarrow{}$ Infograph $\xrightarrow{}$  Sketch  $\xrightarrow{}$ Painting. \\
\emph{Office-Caltech}~\cite{gong2012geodesic} includes 10 categories shared by Office-31~\cite{saenko2010adapting} and Caltech-256~\cite{griffin2007caltech} datasets.
Office-31 dataset contains three domains: DSLR, Amazon and WebCam.
We consider a continual domain adaptation task of DSLR $\xrightarrow{}$  Amazon $\xrightarrow{}$  WebCam $\xrightarrow{}$  Caltech.

\subsubsection{Competing Methods} We compare GRCL  with five alternatives, including (1) DANN~\cite{ganin2016domain}, a classic domain adversarial training based method; (2) MCD~\cite{saito2018maximum}, maximizing the classifier discrepancy to reduce domain gap; (3) DADA~\cite{peng2019domain}, disentangling the domain-specific features from category identity; (4) CUA~\cite{bobu2018adapting}, adopting an adversarial training based method ADDA~\cite{tzeng2017adversarial} to reduce the domain shift and a sample replay loss to avoid forgetting; (5) GRA, replacing the contrastive loss~\ref{eq:2} in GRCL with adversary loss in ADDA~\cite{tzeng2017adversarial}.

\subsubsection{Implementation Details} 
For fair comparison, we adopt LeNet-5~\cite{lecun1998gradient} on Digits, ResNet-50~\cite{he2016deep} on DomainNet, and ResNet-18~\cite{he2016deep} on Office-Caltech.
The number of training and testing images is identical across different domains.
For contrastive learning, we set batch size to be 256, feature update momentum to be  $m=0.5$ in Eq.~\ref{eq:momentum_update}, number of negatives to be 1024 and training schedule to be 240 epochs.
The MLP head uses a hidden dimension of $2048$.
Following~\cite{wu2018unsupervised,he2019momentum}, the temperature $\tau$ in Eq.\ref{eq:2} is $0.07$.
For data augmentation, we use random color jittering, Gaussian blur and random horizontal flip.
To ensure the discriminative ability of features in domain-episodic memories, the image samples and pseudo-labels are generated by clustering with top-1024 confidence. 
For methods using memory, CUA, GRA and GRCL use exactly the same size of domain-episodic memory and $k$-means algorithm to generate pseudo labels.

\subsection{Comparison with Competing Methods}

Table~\ref{tab:t1} summarizes the detailed results for competing methods and GRCL on three continual DA benchmarks.
Each entry in Table~\ref{tab:t1} represents the mean value and standard deviation which are computed by five runs in corresponding experiments. The larger average accuracy (ACC) reflects the better performance of our model in continual DA and larger backward transfer (BWT) reflects better ability to overcome \emph{catastrophic forgetting}.
After the model has been adapted to the final target domain, we report the ACC and BWT over the whole sequential domains.

As shown in Fig.\ref{tab:t1}, GRCL consistently achieves better ACC across three benchmarks, suggesting that the model trained by GRCL owns the best generalization capability across different domains.
Unsurprisingly, most methods exhibit lower negative BWT, as \emph{catastrophic forgetting} exists. 
The methods using memory (CUA, GRA, GRCL) perform better than other methods without memories (DANN, MCD, DADA) by $2\%-5\%$ on ACC and $3\%-5\%$ on BWT. The great improvement on BWT by memory-based methods highlights the importance of memory in the continual DA to overcome \emph{catastrophic forgetting}.

Among the memory-based methods, GRA and GRCL leverage \emph{memorization constraint} while CUA uses replay buffer and multitask loss, \emph{i.e.,} 
\begin{equation} \label{eq:contrast-da-multi-task}
    \begin{aligned}
    \mathcal{L} & = \mathcal{L}_{\mathbf{q}}(\theta_t, \mathcal{B}_t) + \lambda_1 \mathcal{L}_{ce}(\theta_t, \mathcal{D}_s) + \lambda_2 \mathcal{L}_{ce}(\theta_t, \mathcal{M}).
    \end{aligned}
\end{equation}
GRA and GRCL achieve significantly better BWT  on three benchmarks, suggesting the effectiveness of gradient constraints for combating \emph{catastrophic forgetting}.
GRCL consistently achieves better ACC than GRA across all benchmarks.
It is because that GRCL utilizes all the samples from domain memory (cached the samples from all previously observed domains) in contrastive loss to bridge the domain gap, while GRA only uses source domain and current target domain in adversarial loss to learn domain-invariant features.

Fig.\ref{fig:fig3} depicts the evolution of classification accuracy on the first target domain as more domains are adapted to. The accuracy of the first target domain significantly drops in memory-free methods (DANN, MCD, CUA) but the accuracy still maintains in memory-based methods.
GRCL consistently exhibits minimal forgetting and even positive backward transfer on \textit{Office-Caltech} benchmark.

\subsection{Ablation Study}
We analyze the effectiveness of the individual component of GRCL in Table~\ref{tab:ablation_study}. \emph{Src.}: the model is trained on the source domain and then tested on different target domains. \emph{Crt.+Src.}: the model is formulated as a multitask problem but without supervision of classification loss on old target domains (Eq.~\ref{eq:contrast-da-multi-task}, $\lambda_2=0$). \emph{Crt.+Src.+Mem.}: the model is formulated as \emph{Crt.+Src.} plus supervision on domain-episodic memories (Eq.~\ref{eq:contrast-da-multi-task}). \emph{Crt.+SDC}: the model is formulated as a contrastive learning problem (Eq.~\ref{eq:2}) regularized by \emph{Source Discriminative Constraint} (Eq.~\ref{eq:4}). \emph{Crt.+SDC+TMC(GRCL)}: the model is formulated as a contrastive learning problem (Eq.~\ref{eq:2}) regularized by \emph{Source Discriminative Constraint} (Eq.~\ref{eq:4}) and \emph{Target Memorization Constraint} (Eq.~\ref{eq:target_gradient}).

\begin{table}[h]
\renewcommand\arraystretch{1.1}
\centering
\scriptsize
% \footnotesize
\begin{tabular}{@{}lcc|cc@{}}
\toprule
\footnotesize{Methods}                                                                     & \multicolumn{2}{c}{\footnotesize{\textit{DomainNet}}}                                     & \multicolumn{2}{c}{\footnotesize{\textit{Office-Caltech}}} \\ \midrule
                                    & ACC                               & BWT                                    & ACC                                  & BWT                                         \\ % \cmidrule(l){2-3} \cmidrule(l){4-5}
                                    \hline
\emph{Src.}                              & 29.18 \scriptsize $\pm$ 0.05        & -                      & 80.03 \scriptsize $\pm$ 0.05        & -          \\ \hline
\multicolumn{5}{l}{\emph{Multitask Training}}   \\ \hline
\emph{Crt.+Src.}         & 31.53  \scriptsize $\pm$ 0.15        & -8.27  \scriptsize $\pm$ 0.07          & 82.57 \scriptsize $\pm$ 0.03        & -7.46    \scriptsize $\pm$ 0.07     \\
\scriptsize{\emph{Crt.+Src.+Mem.}}         & 35.23  \scriptsize $\pm$ 0.13        & -4.03  \scriptsize $\pm$ 0.11          & 85.37 \scriptsize $\pm$ 0.03        & -3.16    \scriptsize $\pm$ 0.06     \\ \hline
\multicolumn{5}{l}{\emph{Gradient Reguarlized}}   \\ \hline
\emph{Crt.+SDC}         & 33.98  \scriptsize $\pm$ 0.13        & -7.65  \scriptsize $\pm$ 0.05          & 84.75 \scriptsize $\pm$ 0.06        & -6.88    \scriptsize $\pm$ 0.04     \\
% \emph{Crt.+TMC}         & 35.87  \scriptsize $\pm$ 0.13        & -1.27  \scriptsize $\pm$ 0.07          & 86.02 \scriptsize $\pm$ 0.08        & -0.85    \scriptsize $\pm$ 0.03     \\
\emph{Crt.+SDC+TMC}               & 37.74 \scriptsize $\pm$ 0.13        & -0.67  \scriptsize $\pm$ 0.12           & 87.23  \scriptsize $\pm$ 0.06       & 0.05      \scriptsize $\pm$ 0.02            \\ \bottomrule
\end{tabular}
\caption{Ablation studies of GRCL on individual components. }
\label{tab:ablation_study}
\end{table}

\subsubsection{Importance of Contrastive Learning} \label{a:ablation-memory-size}

\begin{table}[h]
        \centering
            \begin{tabular}{ccccc}
            \toprule
            memory size             & 256        & 512       & 1024      & 2048 \\ \midrule
            \textit{Digits}         & 83.00       & 84.12      &  85.34    & 85.41     \\ 
            \textit{DomainNet}      & 33.28      & 35.75     &  37.74     & 37.83      \\
            \bottomrule
            \end{tabular}
        \caption{ACC as a function of memory size.}
        \label{tab:t2-1}
\end{table}

\begin{table}[h]
        \centering
             \begin{tabular}{ccccc}
             \toprule
             training epoch          & 120        & 180             & 240         & 300 \\ \midrule
             \textit{Digits}         & 80.10       & 83.46           & 85.34       & 85.38     \\ 
             \textit{DomainNet}      & 34.80       & 36.50            & 37.74        & 38.16         \\
             \textit{Office-Caltech} & 80.93      & 84.70            & 87.23       & 87.28     \\ \bottomrule
            \end{tabular}
        \caption{ACC as a function of training epoch.}
        \label{tab:t3-1}
\end{table}

Contrastive learning aims to bridge the gap between different domains. As is shown in Table~\ref{tab:ablation_study}, \emph{Crt.+Src.} outperforms \emph{Src.} by about $2.5\%$, which shows contrastive learning can bridge the domain gap by attracting visually similar samples. The BWT in \emph{Crt.+Src.} is higher than other existing memory-free method, \emph{i.e.}, DANN, MCD and DADA, which indicates contrastive loss has the some ability to overcome \emph{catastrophic forgetting}. We attribute this to use memory features in $\mathcal{B}_t$ for contrastive learning while no pseudo-labels are involved. Table~\ref{tab:t2-1} and Table~\ref{tab:t3-1} shows the ACC of GRCL with various sizes of domain-episodic memories and different training epochs. Because contrastive learning naturally benefits from larger memory banks and longer training schedules~\cite{chen2020simple}, GRCL gets consistent improvements with their conclusions.

\subsubsection{Importance of \emph{Source Discriminative Constraint}}
\emph{Source Discriminative Constraint} aims to restore the discriminative ability of features in the source domain and to exploit such discriminative ability to better adapt the model to the target domains. Comparing with \emph{Crt.+Src.} and \emph{Crt.+SDC}, we can see the ACC improves ($\approx 2.2\%$) more than BWT does ($\approx 0.8\%$). The result shows that the improvement of ACC by \emph{Source Discriminative Constraint} mainly results from the model's better adaptation ability to new target domains instead of overcoming \emph{catastrophic forgetting}. Therefore, we conclude \emph{Source Discriminative Constraint} which keeps the discriminative ability of the features in the source domain can benefit domain adaptation to the target domains.

\subsubsection{Importance of \emph{Target Memorization Constraint}}
\emph{Target Memorization Constraint} aims to remember knowledge in old target domains when the model adapts to a new target domain. Comparing with \emph{Crt.+SDC} and \emph{Crt.+SDC+TMC} in Table~\ref{tab:ablation_study}, we can see BWT improves by around $6\%$ and ACC improves by around $3.5\%$. Therefore, we conclude that ACC improvements are mainly from overcoming \emph{catastrophic forgetting} by \emph{Target Memorization Constraint}. Comparing with \emph{Crt.+Src.+Mem.}, \emph{Crt.+SDC+TMC} outperforms ACC by over $2\%$ and BWT by about $4\%$. The improvement on BWT verifies that gradient constraints learning is more effective to restore old knowledge than multitask learning.

\begin{figure}[h]
    \centering
    \includegraphics[width=1\linewidth]{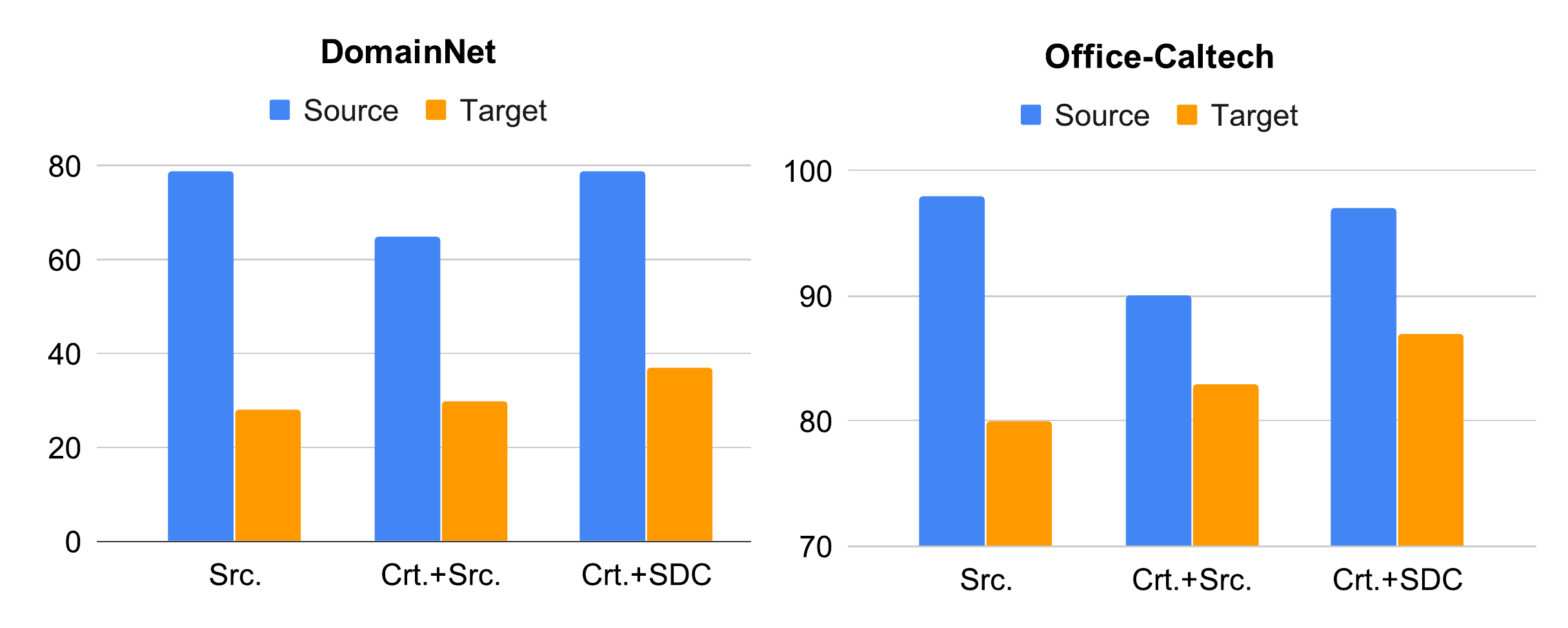}
    \caption{Comparison among \emph{Src.}, \emph{Crt.+Src.} and \emph{Crt.+SDC}. The performance on target represents the averaged accuracy over all different target domains.}
    \label{fig:source-target-acc}
\end{figure}

\subsubsection{Effectiveness of \emph{Source Discriminative Constraint} on Conventional UDA} 
We show the proposed \emph{Source Discriminatve Constraint} outperforms the popular multitask learning in conventional UDA, \emph{i.e.} not Continual DA. We compare three different methods: (1) \emph{Src.}; (2) \emph{Crt.+Src.}; (3) \emph{Crt.+SDC}.
The $\lambda_1$ in Eq.\ref{eq:contrast-da-multi-task} uses the best value obtained via grid search and $\lambda_2=0$. 
The SynNum, Clipart and DSLR are used as the source domain for Digits, DomainNet and Office-Caltech dataset respectively. Rather than ACC and BWT, we evaluate the performance by a different metric for UDA.
We report the averaged classification accuracy on adapting the model from the source domain to the different target domains.
As shown in Fig.\ref{fig:source-target-acc}, 
\emph{Src.} performs well on the source domain, but worse on the target domain, due to the domain gaps.
\emph{Crt.+Src.} improves the performance on the target domain but has a significant adverse effect on the performance on the source domain. \emph{Crt.+SDC} can improve the performance even greater than \emph{Crt.+Src.} on the target domain while maintaining the accuracy on the source domain simultaneously. Therefore, we conclude that because of the \emph{source discriminative constraint}, the discriminative ability of source features is maintained and can further benefit adaptation to the target domain in UDA. 

\section{Conclusion}
This work studies the problem of continual DA, which is one major challenge in the deployment of deep learning models.
We propose Gradient Regularized Contrastive Learning (GRCL) to jointly learn both discriminative and domain-invariant representations.
At the core of our method, gradient regularization maintains the discriminative ability of feature learned by contrastive loss and overcomes \emph{catastrophic forgetting} in the continual adaptation process.
Our experiments demonstrate the competitive performance of GRCL against the state-of-the-art.

% \clearpage
\bibliography{refs}

\end{document}